# Detection of Face Mask using Convolutional Neural Network

Riya Chiragkumar Shah
Department of Computer Science and Engineering
Nirma University
S G highway, Ahmedabad – 382481,
20MCED11@nirmauni.ac.in

Rutva Jignesh Shah
Department of Computer Science and Engineering
Nirma University
S G highway, Ahmedabad – 382481
20MCED12@nirmauni.ac.in

*Abstract*— **In the recent times, the Coronaviruses that are a big family of different viruses have become very common, contagious and dangerous to the whole human kind. It spreads human to human by exhaling the infection breath, which leaves droplets of the virus on different surface which is then inhaled by other person and catches the infection too. So it has become very important to protect ourselves and the people around us from this situation. We can take precautions such as social distancing, washing hands every two hours, using sanitizer, maintaining social distance and the most important wearing a mask. Public use of wearing a masks has become very common everywhere in the whole world now. From that the most affected and devastating condition is of India due to its extreme population in small area. This paper proposes a method to detect the face mask is put on or not for offices, or any other work place with a lot of people coming to work. We have used convolutional neural network for the same. The model is trained on a real world dataset and tested with live video streaming with a good accuracy. Further the accuracy of the model with different hyper parameters and multiple people at different distance and location of the frame is done.**

## I. Introduction

Public use of face masks has been common in China and other nations in the world since the beginning of the new coronavirus disease outbreak. We now know from recent studies that a significant portion of individuals with coronavirus lack symptoms ("asymptomatic") and that even those who eventually develop symptoms ("pre-symptomatic") can transmit the virus to others before showing symptoms, according to the advisory published by the Health Centre. "This means that the virus can spread between people interacting in close proximity — for example, speaking, coughing, or sneezing — even if those people are not exhibiting symptoms". The recent information also gives trace of a new strain of corona virus, the mutant corona virus which, in which the virus has changed its structure and become mutant. The new strain is not even able to detect using the RT-PCR test we use now. So it is inevitable for the people of an overpopulated country like India to wear masks and let the work go on. Nobody can keep an eye on every person coming in the work space is wearing a mask or not. So the need of Face mask detection arose. The model in this paper uses the Convolutional Neural Network. It is a deep neural network model used for analyzing any visual imagery. It takes the image data as input, captures all the data, and send to the layers of neurons. It has a fully connected layer, which processes the final output that represents the prediction about the image. The Convolutional neural network model used here is the MobileNetV2 architecture. MobileNet model is a network model using depth wise separable convolution as its basic unit. Its depth wise separable convolution has two layers: depth wise convolution and point convolution [1] . It is based on an inverted residual structure where the residual connections are between the bottleneck layers. The intermediate expansion layer uses lightweight depth wise convolutions to filter features as a source of non-linearity. As a whole, the architecture of MobileNetV2 contains the initial fully convolution layer with 32 filters, followed by 19 residual bottleneck layers. Figure 1 shows the framework of MobileNetV2 which is used in the model discussed in this paper.

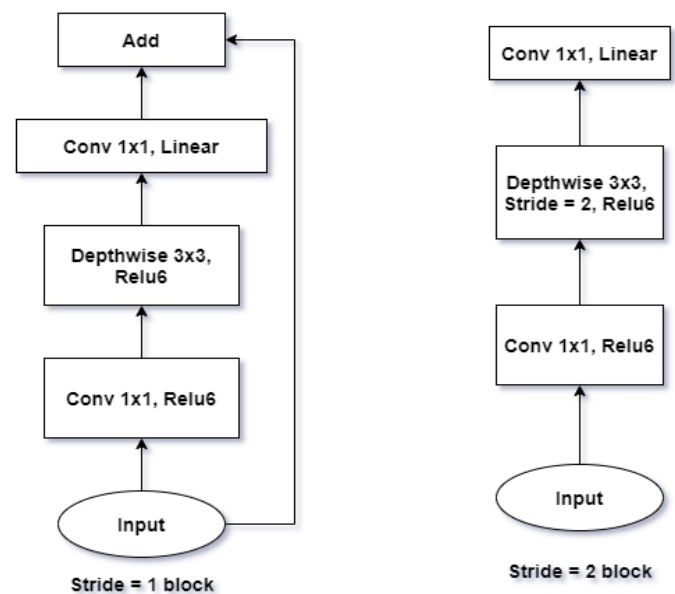

*Figure1. MobileNetV2*

Further the different hyper parameters are tried for the model. The hyper parameters tried are learning rate, it is a tuning parameter that is used in optimization models which determines the step size of the model and helps to reduces the loss function. It is a very important hyper parameter as it results in either convergence or overshoots the model. The other hyper parameters used are batch size, epochs etc. The model has used OpenCV to fulfil the purpose of using the video stream for capturing the frames in the video stream.

## II. RELATED WORK

In [3] they have proposed a pre-trained MobileNet with a global pooling block for face mask detection. The pre-prepared MobileNet takes a shading picture and creates a multi-dimensional component map. The worldwide pooling block that has been used in the proposed model changes the element map into an element vector of 64 highlights. At long last, the softmax layer performs paired order utilizing the 64 highlights. We have assessed our proposed model on two openly accessible datasets. Our proposed model has accomplished 99% and 100% exactness on DS1 and DS2 separately. The worldwide pooling block that has been utilized in the proposed model dodges overfitting the model. Further, the proposed model beats existing models in the quantity of boundaries just as preparing time. But this model cannot detect face mask for multiple faces at a time. In [5] paper utilizes a proficient and strong item location calculation to naturally identify the appearances with veils or without covers, making the plague avoidance work more clever. In particular, they gathered a broad data set of 9886 pictures of individuals with and without face covers and physically named them, at that point use multi-scale preparing and picture mistake techniques to improve YOLOv3, an article recognition calculation, to consequently distinguish whether a face is wearing a veil. Our analysis results show that the mean Average Precision (mAP) of the improved YOLOv3 calculation model came to 86.3%. This work can viably and naturally distinguish whether individuals are wearing veils, which decreases the pressing factor of conveying HR for checking covers openly puts and has high functional application esteem.

## III. PROPOSED SYSTEM

The model proposed here is designed and modeled using python libraries namely Tensorflow, Keras and OpenCV. The model we used is the MobileNetV2 of convolurional neural network. The method of using MobileNetV2 is called using Transfer Learning. Transfer learning is using some pre trained model to train your present model and get the prediction which saves time and makes using training the different models easy. We tune the model with the hyper parameters : learning rate, number of epochs and batch size. The model is trained with a dataset of images with two class, with mask and without mask. The dataset has 993 images of with mask class and 1918 images of without mask class.

(i) Training the model with the taken dataset.
(ii) Deploying the model

In the paper we have developed a model using the above mentioned libraries. We have tested the model for different conditions with different hyper parameters, for which the results are mentioned in the next section. First we feed the dataset in the model, run the training program, which trains the model on the given dataset. Then we run the detection program, which turns on the video stream, captures the frames continuously from the video stream with an anchor box using object detection process. This is passed through the MobileNetV2 model layers which classifies the image as with or without mask. If the person is wearing a mask, a green anchor box is displayed and red if not wearing a mask with the accuracy for the same tagged on the anchor box. Figure2 shows the flow of the Face Mask Detection model used in this paper.

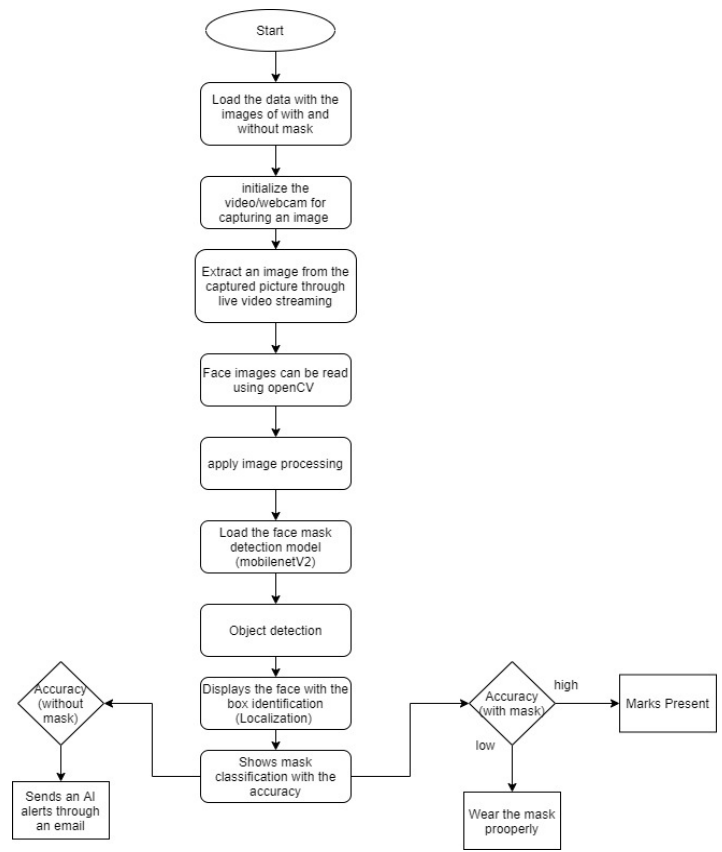

*Figure2. Flow of Face Mask Detection Model*

The face mask recognition system uses AI technology to detect the person with or without a mask. It can be connected with any surveillance system installed at your premise. The authorities or admin can check the person through the system to confirm their identity. The system sends an alert message to the authorized person if someone has entered the premises without a face mask. The accuracy rate of detecting a person with a face mask is 95-97% depending on the digital capabilities. The data has been transferred and stored automatically in the system to enable reports whenever you want.

## IV. RESULTS

We have tested the model for different scenarios, below mentioned is the table with the results of those scenarios with number of epochs 20 and batch size 32 constant for all the three situations. We have used Average Pooling for capturing smooth image. Table 1 shows the results of comparison of different hyper parameters and situations

| Model | Learning rate | With mask distance | Without mask distance | Blur image quality | Multiple people capturing |
|---|---|---|---|---|---|
| 1 | 1e-4 | 161 cm | 190 cm | Good | 4 people |
| 2 | 1e-3 | 155 cm | 187 cm | Average | 3 people |
| 3 | 1e-2 | 146 cm | 179 cm | Bad | 3 people |

*Table 1. Result Comparison Table*

According to the above results the first model is the best compared to all the models. The plot of the best model from our research is shown below. Its shows the plot for training loss, validation loss, training accuracy and validation accuracy for Number of epochs versus loss or accuracy. It is evident from the plot that as the number of epochs increases the training and validation accuracy increases and the training and validation accuracy decreases. And also the validation accuracy is higher than the training accuracy which proves that the model is not suffering through overfitting. Figure 3 shows the plot for number of epochs versus loss or accuracy.

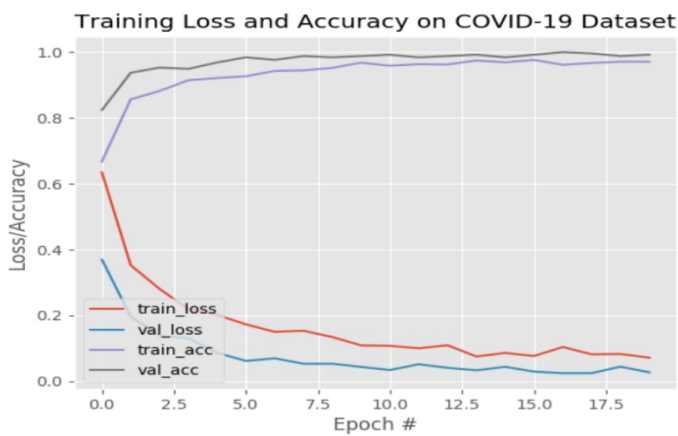

*Figure3. Graph of Number of epochs vs loss or accuracy.*

## V. FUTURE WORK

The present model proposed gives great accuracy for single face with and without mask. For multiple faces also it gives quite good accuracy. It works easily on any mobile device just by switching on the video stream, with no external hardware requirement. Further we will work for improving the accuracy for multiple face mask detection, to classify the faces into three categories that is, With mask, without mask, Improper mask instead of just the two with and without mask class by adding datasets with images of people wearing masks not covering their noses properly and also to detect the masked face using the FaceNet model of Convolutional Neural Network like in [4] so as to further improve our model and add marking attendance feature in it by detecting the face even when the mask is on.

## VI. CONCLUSION

To moderate the spread of the COVID-19 pandemic, measures should be taken. We have demonstrated a facemask detector using Convolutional Neural Network and move learning techniques in neural organizations. To train, validate and test the model, we utilized the dataset that consisted of 993 masked faces pictures and 1918 exposed faces pictures. These pictures were taken from different assets like Kaggle and RMFD datasets. The model was induced on pictures and live video transfers. To choose a base model, we assessed the measurements like precision, accuracy, and recall and chose MobileNetV2 architecture with the best exhibition having 99% precision and 99% recall. It is additionally computationally efficient using MobileNetV2 which makes it simpler to introduce the model to inserted frameworks. This face mask detector can be sent in numerous regions like shopping centers, air terminals and other substantial traffic places to screen people in general and to dodge the spread of the infection by checking who is following essential rules and who isn't.